\documentclass[runningheads]{llncs}
\usepackage{graphicx}
\usepackage{todonotes}
\usepackage{url}
\usepackage{listings}
\usepackage{subcaption}
\usepackage{float}

\begin{document}
\title{OntoChat: a Framework for Conversational Ontology Engineering using Language Models}
\titlerunning{OntoChat: a Framework for Conversational Ontology Engineering}
%
\author{Bohui Zhang\inst{1}\orcidID{0000-0001-5430-1624} \and
Valentina Anita Carriero\inst{2}\orcidID{0000-0003-1427-3723} \and
Katrin Schreiberhuber\inst{3}\orcidID{0000-0003-1815-8167} \and
Stefani Tsaneva\inst{3}\orcidID{0000-0002-0895-6379} \and
Lucía Sánchez González\inst{4}\orcidID{0000-0002-1685-4700} \and
Jongmo Kim\inst{1}\orcidID{0000-0002-4984-1674} \and
Jacopo de Berardinis\inst{1}\orcidID{0000-0001-6770-1969}}
\authorrunning{Zhang et al.}
%
\institute{Department of Informatics, King's College London, London, UK \and
Cefriel -- Politecnico di Milano, Milan, Italy \and
Vienna University of Economics and Business, Vienna, Austria\and
Ontology Engineering Group, Universidad Politécnica de Madrid, Spain\\
\email{bohui.zhang@kcl.ac.uk}}
\maketitle              

\begin{abstract}
Ontology engineering (OE) in large projects poses a number of challenges arising from the heterogeneous backgrounds of the various stakeholders, domain experts, and their complex interactions with ontology designers.
This multi-party interaction often creates systematic ambiguities and biases from the elicitation of ontology requirements, which directly affect the design, evaluation and may jeopardise the target reuse.
Meanwhile, current OE methodologies strongly rely on manual activities (e.g., interviews, discussion pages).
After collecting evidence on the most crucial OE activities, we introduce \textbf{OntoChat}, a framework for conversational ontology engineering that supports requirement elicitation, analysis, and testing.
By interacting with a conversational agent, users can steer the creation of user stories and the extraction of competency questions, while receiving computational support to analyse the overall requirements and test early versions of the resulting ontologies.
We evaluate OntoChat by replicating the engineering of the Music Meta Ontology, and collecting preliminary metrics on the effectiveness of each component from users.
We release all code at \url{https://github.com/King-s-Knowledge-Graph-Lab/OntoChat}.

\end{abstract}
\section{Introduction}\label{sec:introduction}

Ontology Engineering (OE) encompasses a number of activities defining a collaborative effort to design, evaluate, and reuse ontologies of general or domain-specific purposes~\cite{kendall2019ontology,simperl2014collaborative}.
Depending on the intended reuse by one or more stakeholders, the process starts with ontology designers eliciting requirements from the former, while interacting with experts to validate their formal understanding of the domain.
Although the way OE activities are organised may vary depending on the methodology adopted \cite{Blomqvist2010,suarez2012neon}, these interactions are frequent and iterative throughout the development cycle.
In large projects, ontology designers engage in a number of manual activities and interact with multiple parties to make sure requirements are collected consistently and comprehensively, seeking clarification from domain experts to formalise and test them~\cite{polifonia2021report}.
This multi-party interaction may require substantial resources and create systematic ambiguities and biases arising from potentially conflicting or ill-formulated requirements \cite{polifonia2023pon}.
Not only does this affect the ontology design and evaluation, but it may also jeopardise their reuse by the same stakeholders.
In other cases, a blurry line exists among stakeholders, domain experts, and ontology designers, which is common to multidisciplinary and highly collaborative projects, such as Wikidata~\cite{vrandecic-krotzsch-2014-wikidata}.

OE has long faced arguments about its technical challenges and costs.
Despite the development of collaborative methodologies (c.f. Section~\ref{sec:related-work}), the field still faces concerns about positively impacting the liveliness, evolution, and reusability of ontologies \cite{mateiu2023ontology}.
Meanwhile, Large Language Models (LLMs) have received increasing attention in the Semantic Web due to their language understanding capabilities.
These provide generalisation across various tasks (e.g., question answering, text summarisation), making it possible to effectively generate, process, and annotate text to address knowledge engineering tasks \cite{allen2023knowledge,zhang2023llmke}.

Here, we hypothesise that LLMs can assist and facilitate OE activities to implement conversational workflows and prompt-driven functionalities that can reduce the complex interactions between domain experts and ontology designers, while accelerating the analysis and formalisation of requirements.
To confirm our motivations and collect insights for the design of such a framework, we put forth the following research questions.

\begin{itemize}
    \item \textbf{RQ1}: Which ontology engineering activities are the most in need of computational support?
    \item \textbf{RQ2}: How can LLMs enable a conversational ontology engineering framework to support these activities?
\end{itemize}

To address RQ1, we conducted a survey asking ontology engineers to rate OE activities in relation to their difficulty and the manual effort required.
We reuse the insights collected from the survey to implement a conversational workflow where different categories of users interact independently with the system and produce intermediate artefacts and documentation throughout the various stages of the OE process (RQ2).
In sum, this work contributed:

\begin{itemize}
    \item An investigation of which OE activities are most in need of computational support, gathered from a survey involving participants with OE expertise.
    \item \textbf{OntoChat}, a conversational framework for ontology engineering providing support for (i) \textit{requirement elicitation} (user story creation, competency question extraction), (ii) \textit{analysis} (competency question verification, reduction, and clustering), and for (iii) \textit{testing} preliminary versions of an ontology.
    \item A preliminary evaluation of an online prototype of OntoChat where participants were asked to replicate the OE of the Music Meta ontology \cite{de2023music} and measure the effectiveness of their outcomes and interactions.
\end{itemize}

\section{Related work}\label{sec:related-work}


Various OE methodologies have been proposed over the years, with their focus shifting towards collaborative approaches \cite{simperl2014collaborative}.
Early works include METHONTOLOGY \cite{fernandez1997methontology}, based on requirements elicitation from \cite{uschold1995towards} and providing support for conceptualisation, implementation, and maintenance steps; and DILIGENT \cite{pinto2004diligent} which also account for the involvement of different stakeholders.
More recent work incorporated Agile principles to support iterative ontology development, such as NEON \cite{suarez2012neon}, SAMOD \cite{peroni2016samod}, and \textit{eXtreme Design} (XD) \cite{Blomqvist2010,Blomqvist2016}.
The latter also provides guidelines for requirement elicitation and is strongly test-based.


In \cite{polifonia2023pon}, ontology requirements are collected from \emph{customers} in the form of \emph{user stories}.
A story\footnote{Examples of user stories at \url{https://github.com/polifonia-project/stories}} contains three main components:
the \textit{persona} portrays a typical user, including their name, occupation, skills, and interests;
the \textit{goal} captures the persona's aims in the story;
the \textit{scenario} describes how the persona's goals are currently addressed, to contextualise the gap with the resource being developed.
Through the collection of stories, ontology requirements can then be defined by extracting competency questions (CQs) -- the natural language counterpart of structured queries that the resulting knowledge graph (KG) should answer~\cite{GruningerFox94}.

CQs are central to OE.
They facilitate the collection of requirements, drive the implementation of the ontologies (e.g., in XD, they are mapped to ontology design patterns) and are used for testing \cite{blomqvist2012ontology}.
Nonetheless, despite the level of experience in a particular OE methodology, bottlenecks, ambiguities, and domain jargon can still hamper progress from the requirement collection stage \cite{polifonia2021report}.
This especially happens when several stakeholders and domain experts are involved.
To mitigate this issue, \cite{polifonia2023pon} introduced IDEA -- a tool supporting the iterative elicitation and improvement of CQs via NLP methods.

Recently, LLMs have been applied to support and augment knowledge engineering tasks.
Applications range from representing domain knowledge and generating examples of classes and relations to providing explanations and recommendations on ontologies after verbalising them into plain text \cite{meyer2023llm}.
For ontology engineering, LLM-based approaches have shown promising results for ontology matching and alignment \cite{qiang2023agent,he2023exploring,hertling2023olala}.
Other studies have focused on ontology construction and learning from text, using LLMs to suggest relevant subconcepts \cite{funk2023towards} and to automatically extract and structuring knowledge \cite{babaei2023llms4ol}.
In the context of requirement elicitation, new methods were contributed for retrofitting CQs from ontologies to promote reuse \cite{alharbi2023experiment} and extracting CQs directly from KGs \cite{polifonia2023revont}.
Overall, these works demonstrated great potential for LLM-driven knowledge engineering, but also acknowledged significant issues such as hallucination, poor non-linguistic reasoning, and the high cost of fine-tuning.

\section{OntoChat: towards conversational ontology engineering}\label{sec:ontochat}

\begin{figure}[ht]
    \centering
    \includegraphics[width=\linewidth]{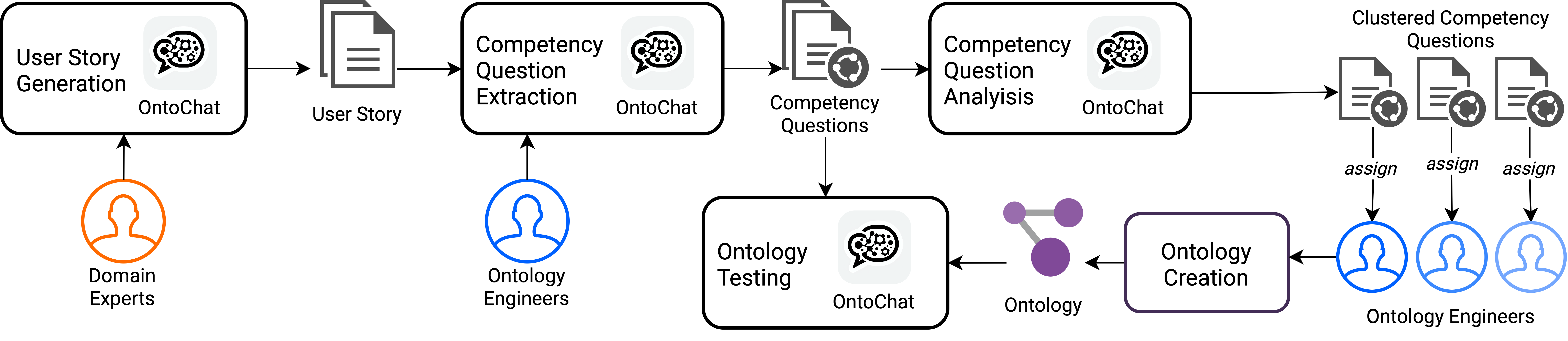}
    \caption{Illustration of the workflow alongside the main features in OntoChat.}
    \label{fig:ontochat-workflow}
\end{figure}

To address RQ1 and inform the design of our framework, we conducted an online survey asking participants to rank OE activities for complexity and need of computational support.
The survey was conducted via Google Forms (without recording any personal data from participants) and was distributed to Semantic Web practitioners.
Participants were asked to quantify the agreement of statements on a 5-point Likert scale, with each statement expressing an OE activity.

We gathered responses from $N=23$ participants with various level of experience and familiarity with OE methodologies (primarily Neon, Ontology 101, Linked Open Terms).
Results are detailed in Appx~\ref{sec:oe-survey}.
Given the size of our sample, we decided to rely on strong evidence ($\geq75\%$ positive responses using 4-5 scores), finding that the most demanding OE tasks in need of computational support are: the collection of ontology requirements ($86.4\%$), the extraction of CQs from textual ontology requirements ($81.8\%)$, the analysis of ontology requirements ($77.3\%$), and ontology testing ($81.8\%$).

Based on these findings, and building upon the IDEA approach \cite{polifonia2023pon}, we designed OntoChat to support such OE activities.
The workflow, illustrated in Fig.~\ref{fig:ontochat-workflow}, leverages LLMs as knowledge elicitators to reduce the demand and complexity of the multi-party interactions with ontology designers, and aims at accelerating OE tasks.
To collect requirements, the process starts with stakeholders and domain experts co-creating user stories by interacting with our conversational agent.
Ontology engineers can then extract CQs by iteratively refining the model's recommendations; reduce redundant requirements, and analyse the resulting CQs via clustering.
Finally, OntoChat also allows testing preliminary versions of ontologies via verbalisation and unit prompting.

\subsection{Assisted persona and story creation}\label{ssec:ontochat-story}


To enable the collaborative user story generation OntoChat follows steps below. 

\textit{Step 1: LLM role definition (back-end).} 
The LLM behaviour is primed to emulate a knowledge elicitator aiming to gather information from the user about each user story component -- the persona definition, specification of concrete goals \& addressed scenario, and the provision of data examples (back-end).

\textit{Step 2: Knowledge elicitation (user involvement).} 
The user is guided through a series of questions focusing on a particular story aspect. For instance, to gather insights about the persona's background, OntoChat asks
\textit{``What are the name, occupations, skills, interests of the user?"} 
Whenever the user does not provide enough details or gives partial answers, the LLM continues to elicit additional insights until all necessary information is gathered (c.f. Fig.~\ref{fig:story-elicitaion-partial}, Appx.~\ref{app-userstory-prompts}).

\textit{Step 3: User story generation (back-end)}
The initial user story draft is created following a one-shot learning approach \cite{brown2020language}. The LLM is supplied with a user story example, and is prompted to follow the same structure for generating a \textit{user story draft} summarising the information extracted in the elicitation step.  

\textit{Step 4: Refinement (user involvement).} The user is presented with the user story draft and is encouraged to provide feedback. The refinement stage is iterative and continues until the user no longer requests further changes. 
Possible refinements include the correction of factual inconsistencies,   additions to the user stories, removal of irrelevant details, etc. We provide some exemplary refinement requests in Fig.~\ref{fig:story-refinement} (Appx.~\ref{app-userstory-prompts}).

\subsection{Competency question extraction}\label{ssec:ontochat-cqext}

The main objective of this module is to assist ontology engineers in extracting CQs from user stories.
The procedure is organised as follows.

\textit{Step 1: Instructing the LLM for CQ extraction (back-end).} 
The model is provided with examples of pairs of user story fragments and expected CQs, to align its outputs to the expectations of ontology engineers.

\textit{Step 2: First extraction of CQs (user involvement).}
The ontology engineer is asked to provide a user story for CQ extraction.
The user story may be manually crafted, or obtained from the previous step (see Fig.~\ref{fig:story-example}, Appx.~\ref{app-userstory-prompts}).
As output, OntoChat provides an initial list of CQs (see Fig.~\ref{fig:CQ_first_input}, Appx.~\ref{app-cq-prompts})

\textit{Step 3: Competency question refinement (back-end).} 
The LLM is provided with a series of prompts (hidden from the user) to perform two \textit{refinement steps}.

\begin{itemize}
    \item \textit{Step 3.1: Split not atomic CQs.} If the example data is complex, users may get non-atomic questions from a single example.
    As these usually entail nested requirements, complex CQs need to be split.
    Following a few-shot learning approach, the LLM is asked whether each CQ has a complex form, hence triggering the simplification.
    For example, from the data ``\textit{The musical work Penny Lane has genre/style baroque pop and psychedelic pop.}'', the LLM generated ``\textit{What genres/styles are associated with Penny Lane?}''.
    After this step, the LLM replaces it with two distinct CQs: ``\textit{What genres are associated with Penny Lane?}'' and ``\textit{What styles are associated with Penny Lane?}''.

    \item \textit{Step 3.2: Named entities abstraction.} As an example, the previous CQs replaced the specific genres with the interrogative pronoun ``what'' (genres/styles).
    However, it did not remove the real-world entity ``Penny Lane''. Within this step, the LLMs is prompted to check again the CQs, and, guided by examples, remove possible named entities. This yields abstract CQs like ``\textit{What genres are associated with the musical work?}'' and ``\textit{What styles are associated with the musical work?}''.

\end{itemize}

\textit{Step 4: User confirmation (user involvement).} Finally, OntoChat asks the user whether the number of CQs and their formulation are sound. If not, by leveraging knowledge acquired from previous prompts (see Step 3), the model repeats the refinement steps until the user is satisfied (see Fig.~\ref{fig:CQ_refinement}, Appx.~\ref{app-cq-prompts}).

\subsection{Competency question filtration and analysis}\label{ssec:ontochat-cqan}

As some CQs may be redundant or show negligible semantic variations that are of little interest to ontology engineers, OntoChat provides support for their filtration and analysis.
This is achieved through: \textit{paraphrase identification}, to remove equivalent CQs; and \textit{CQ clustering}, to identify groups of similar requirements.
In \cite{polifonia2023revont}, the former was found to have two benefits: (i) it mitigates the noise and the artefacts introduced in the previous steps; (ii) it reduces the number of CQs that will be presented to ontology engineers.

In contrast to \cite{polifonia2023revont,polifonia2023pon}, which both rely on sentence embeddings and specialised models, this functionality is entirely supported by LLMs, which is motivated by recent findings demonstrating that LLMs possess clustering capabilities~\cite{aharoni-goldberg-2020-unsupervised,zhang-et-al-2023-clusterllm,viswanathan-et-al-2023-few-shot}.
Given a list of CQs, the LLM is asked to remove redundant questions and find meaningful groups of CQs sharing the same thematic focus and intent.
The latter is expected to support ontology designers in understanding requirements and possibly organising their Agile teams (e.g., a team receiving a CQ cluster based on their familiarity with the sub-domain).
In the current version, this step does not require user supervision.

\subsection{Ontology testing support}\label{ssec:ontochat-pretesting}

While the previous functionalities focus on requirement elicitation and analysis, this component provides support for testing preliminary or iterative versions of an ontology.
Ontology testing efforts are often categorised into three methodologies: CQ verification, inference verification, and error provocation.
The first two are concerned with verifying the correct implementation of a requirement, whereas the latter is needed to find cases where the ontology should fail \cite{blomqvist2012ontology}.
These are typically done by formalising CQs into SPARQL queries. 

To test preliminary versions of an ontology, we aim for a SPARQL-free approach to achieve fast CQ verification and inference, while supporting error provocation.
This is achieved in two steps: \textit{ontology verbalisation}, and prompt-driven \textit{CQ unit testing}.
Our verbalisation converts a OWL ontology into plain text by documenting classes, properties, named entities, and their relationships in a descriptional manner.
The method follows a simple algorithmic procedure and assumes that the ontology is well commented to produce an expressive verbalisation.
Then, using the verbalisation, the LLM is prompted to asses the coverage of each CQ by replying \textit{Yes}/\textit{No}.
To prevent prompt leakage and ensure independence in the model's predictions, this is done separately for each CQ.

\subsection{Implementation details}\label{ssec:ontochat-implementation}

OntoChat is implemented in \texttt{Python 3.11} and is released on GitHub\footnote{\url{https://github.com/King-s-Knowledge-Graph-Lab/OntoChat}} (code, prompts, experiments)
under the MIT license.
To facilitate its use and collect user feedback, we implemented an interface prototype using Gradio~\cite{abid2019gradio}.
This can be launched on a local server, also hosted on Hugging Face Spaces\footnote{\url{https://huggingface.co/spaces/b289zhan/OntoChat}}.
The interface has four tabs that wrap all the 
functionalities within the same environment.
The current version uses OpenAI's API and has been evaluated on the GPT-3 family of models (\texttt{gpt-3.5-turbo}, and \texttt{gpt-3.5-turbo-16k} for larger contexts).

\section{Evaluation}\label{sec:evaluation}

We performed a component-based evaluation of OntoChat to measure the effectiveness of each functionality and collect user feedback based on their experiences.
The evaluation was organised to replicate the OE activities of the Music Meta ontology \cite{de2023music}.
It was chosen as a benchmark/testbed for three reasons: it required considerable OE efforts and was already the source of ambiguities in the Polifonia project \cite{polifonia2023d22}; it was complemented by high-quality material (user stories, CQs, documentation, queries, etc.) from \cite{polifonia2023pon} to use as ground truth; the authors had access to a pool of domain experts for evaluation.

\subsection{Experimental methodology}

To ensure each component is evaluated individually by the intended target users, we evaluate the more generative components by collecting feedback on their use from domain experts and ontology engineers through questionnaires.
All questions ask participants to quantify the agreement with the statement made from 1 (strongly disagree) to 5 (strongly agree), with 3 being a neutral response (NR).
No personal information is collected throughout the evaluation.
The ontology testing feature, instead, is evaluated experimentally for accuracy.

\paragraph{User story questionnaire (domain experts).} 
We recruited $N=6$ music experts in the Polifonia project to create user stories summarising their requirements on music metadata using OntoChat.
Our goal is to evaluate the model's success in producing satisfactory user stories. Additionally, we gather insights on OntoChat's usability and its effectiveness in minimising manual effort.

\paragraph{CQ extraction and clustering questionnaires (ontology engineers).}
We recruited $N=8$ ontology engineers to evaluate the model's performance in generating CQs that are consistent with the given story, well-formulated, and consistent with the intended scope of the ontology.
Users are first introduced to the concept of the user story, then asked to familiarise with the \textit{Linka -- Music Knowledge} story\footnote{\url{https://github.com/polifonia-project/stories/tree/main/Linka_Computer_Scientist}}.
Their task is to extract and analyse CQs from the story using OntoChat.

\paragraph{Ontology testing evaluation.}
Given the OWL definition of Music Meta, and the 28 (manually produced) CQs driving its implementation, we evaluate this component as a classification task.
This is done by extending the CQ set with the same number of negative CQs (requirements that are not yet supported by Music Meta).
We expect OntoChat to correctly discriminate between these groups.

\subsection{Preliminary results}


Feedback collected from domain experts confirmed that the user stories generated with OntoChat captured the intended goal and requirements and always provided relevant information (Fig.~\ref{fig:evaluation-stories}).
More than 80\% of participants enjoyed using the tool and found the final stories well-structured and easily understandable.
While users acknowledged the model's ability to improve intermediate drafts through their feedback, only 50\% were satisfied with the example data generated.
Overall, 4/6 experts recognised the tool's potential to accelerate this task (2 NR), and 5 of them would prefer it over manual curation.

From the evaluation with ontology engineers, OntoChat was found to generate CQs that are comprehensive, reflective of the intended ontology scope, and easy to understand (Fig.~\ref{fig:evaluation-cqe}).
However, 2/8 participants noted the extraction of entities outside the story's scope; and only 50\% observed the potential to reduce possible author bias.
While 6/8 participants expressed satisfaction with OntoChat and recognised its time-saving benefits, all agreed it holds promise for streamlining CQ generation, indicating a preference over fully manual creation.

The clustering feature proved advantageous for understanding and organising ontology requirements when compared to full manual inspection (Fig.~\ref{fig:evaluation-cqc}).
Participants found the interaction intuitive (62.5\%) and the resulting clusters expressed meaningful groupings of CQs (87.5\%).
While the feature offers time-saving benefits by providing an aggregated view of ontology requirements, there were indications that it may not fully support comprehensive analysis on its own.

Finally, for ontology testing, we found that OntoChat can correctly classify supported requirements with an accuracy of 87.5\% (P=88\%, R=85.7\%), and often provides explanations and examples to support its prediction (Appx.~\ref{ssec:apx-eval-test}).
\section{Conclusions}\label{sec:conclusions}

This work addresses the challenges of ontology engineering in large collaborative projects by implementing a conversational workflow to streamline the process.
The proposed framework, OntoChat, leverages LLMs to facilitate requirement elicitation, analysis, and ontology testing.
Our preliminary evaluation efforts demonstrate a positive response from domain experts and ontology engineers, indicating potential for accelerating conventional ontology engineering tasks.

Nonetheless, several limitations still exist, notably those inherent to the use of LLMs in specialised domains due to their limited or potentially obsolete knowledge.
Additional challenges include addressing biases in persona creation and enhancing the framework to provide insights into implementation costs and timelines.
This will allow us to measure the amount of user supervision and involvement (e.g., number of interactions with the LLM, specificity of user feedback) during the refinement steps, needed to achieve a reasonable output from OntoChat (e.g., a user story, a list of competency questions), in contrast to full manual curation.
Future work will focus on addressing these challenges, while enhancing the generation of examples in user stories, refining named entity scope in competency question creation, and broadening analysis support.

\subsubsection{Acknowledgements} This project has received funding from the European Union's Horizon 2020 research and innovation programme under grant agreement No 101004746. This work was partly funded by the HE project MuseIT, which has been co-founded by the European Union under the Grant Agreement No 101061441.

%
%
%
\bibliographystyle{splncs04}
\bibliography{main}

\begin{thebibliography}{10}
\providecommand{\url}[1]{\texttt{#1}}
\providecommand{\urlprefix}{URL }
\providecommand{\doi}[1]{https://doi.org/#1}

\bibitem{abid2019gradio}
Abid, A., Abdalla, A., Abid, A., Khan, D., Alfozan, A., Zou, J.: {Gradio: Hassle-Free
Sharing and Testing of ML Models in the Wild}. arXiv preprint arXiv:1906.02569  (2019)

\bibitem{aharoni-goldberg-2020-unsupervised}
Aharoni, R., Goldberg, Y.: Unsupervised domain clusters in pretrained language models. In: Jurafsky, D., Chai, J., Schluter, N., Tetreault, J. (eds.) Proceedings of the 58th Annual Meeting of the Association for Computational Linguistics. pp. 7747--7763. Association for Computational Linguistics, Online (Jul 2020). \doi{10.18653/v1/2020.acl-main.692}

\bibitem{alharbi2023experiment}
Alharbi, R., Tamma, V., Grasso, F., Payne, T.: {An Experiment in Retrofitting Competency Questions for Existing Ontologies}. arXiv preprint arXiv:2311.05662  (2023)

\bibitem{allen2023knowledge}
Allen, B.P., Stork, L., Groth, P.: {Knowledge Engineering using Large Language Models}. arXiv preprint arXiv:2310.00637  (2023)

\bibitem{babaei2023llms4ol}
Babaei~Giglou, H., D’Souza, J., Auer, S.: {LLMs4OL: Large language models for ontology learning}. In: International Semantic Web Conference. pp. 408--427. Springer (2023)

\bibitem{polifonia2023pon}
de~Berardinis, J., Carriero, V.A., Jain, N., Lazzari, N., Meroño-Peñuela, A., Poltronieri, A., Presutti, V.: {The Polifonia Ontology Network: Building a Semantic Backbone for Musical Heritage}. In: Proceedings of the 22nd International Semantic Web Conference (ISWC) (2023)

\bibitem{de2023music}
de~Berardinis, J., Carriero, V.A., Meroño-Peñuela, A., Poltronieri, A., Presutti, V.: {The Music Meta Ontology: a flexible semantic model for the interoperability of music metadata}. In: Proceedings of the the 24th International Society for Music Information Retrieval Conference (2023)

\bibitem{polifonia2023d22}
de~Berardinis, J., Peñuela, A.M., Jain, N., Poltronieri, A., Lazzari, N., Presutti, V., Rigaux, P., Marzi, E., Graciotti, A., Wigham, M., Tirado, A.M., Gurrieri, M., Carriero, V.A., van Kranenburg, P.: {Ontologies and knowledge graphs of music objects, patterns, and software package – 2nd version}. Tech. rep., European Commission, The Polifonia consortium (2023)

\bibitem{Blomqvist2016}
Blomqvist, E., Hammar, K., Presutti, V.: {Engineering Ontologies with Patterns - The eXtreme Design Methodology}. In: Ontology Engineering with Ontology Design Patterns - Foundations and Applications, Studies on the Semantic Web, vol.~25. IOS Press, Amsterdam (2016). \doi{10.3233/978-1-61499-676-7-23}

\bibitem{Blomqvist2010}
Blomqvist, E., Presutti, V., Daga, E., Gangemi, A.: {Experimenting with eXtreme Design}. In: Knowledge Engineering and Management by the Masses. EKAW 2010. vol.~6317, pp. 120--134. Springer, Berlin, Heidelberg (2010). \doi{10.1007/978-3-642-16438-5\_9}

\bibitem{blomqvist2012ontology}
Blomqvist, E., Seil~Sepour, A., Presutti, V.: {Ontology testing-methodology and tool}. In: Knowledge Engineering and Knowledge Management: 18th International Conference, EKAW 2012, Galway City, Ireland, October 8-12, 2012. Proceedings 18. pp. 216--226. Springer (2012)

\bibitem{polifonia2021report}
Bottini, T., Carriero, V.A., Carvalho, J., Cath{\'{e}}, P., Ciroku, F., Daga, E., Daquino, M., Davy-Rigaux, A., Guillotel-Nothmann, Gurrieri, M., van Kemenade, P., Marzi, E., Mero\~no Pe\~nuelala, A., Mulholland, P., Musumeci, E., Presutti, V., Scharnhorst, A.: {D1.1 Roadmap and pilot requirements 1st version}. Tech. rep., EU Commission, The Polifonia consortium (2021)

\bibitem{brown2020language}
Brown, T., Mann, B., Ryder, N., Subbiah, M., Kaplan, J.D., Dhariwal, P., Neelakantan, A., Shyam, P., Sastry, G., Askell, A., et~al.: {Language models are few-shot learners}. Advances in neural information processing systems  \textbf{33},  1877--1901 (2020)

\bibitem{polifonia2023revont}
Ciroku, F., de~Berardinis, J., Kim, J., Mero{\~n}o-Pe{\~n}uela, A., Presutti, V., Simperl, E.: {RevOnt: Reverse Engineering of Competency Questions from Knowledge Graphs via Language Models}. Manuscript under review  (2024)

\bibitem{fernandez1997methontology}
Fern{\'a}ndez-L{\'o}pez, M., G{\'o}mez-P{\'e}rez, A., Juristo, N.: {METHONTOLOGY: From Ontological Art Towards Ontological Engineering}. In: AAAI Conference on Artificial Intelligence (1997)

\bibitem{funk2023towards}
Funk, M., Hosemann, S., Jung, J.C., Lutz, C.: {Towards Ontology Construction with Language Models}. arXiv preprint arXiv:2309.09898  (2023)

\bibitem{GruningerFox94}
Gruninger, M., Fox, M.S.: {The role of competency questions in enterprise engineering}. In: Benchmarking — Theory and Practice. IFIP Advances in Information and Communication Technology. pp. 83--95. Springer, Boston, MA (1994)

\bibitem{he2023exploring}
He, Y., Chen, J., Dong, H., Horrocks, I.: {Exploring large language models for ontology alignment}. arXiv preprint arXiv:2309.07172  (2023)

\bibitem{hertling2023olala}
Hertling, S., Paulheim, H.: {Olala: Ontology matching with large language models}. In: Proceedings of the 12th Knowledge Capture Conference 2023. pp. 131--139 (2023)

\bibitem{kendall2019ontology}
Kendall, E.F., McGuinness, D.L.: {Ontology engineering}. Morgan \& Claypool Publishers (2019)

\bibitem{mateiu2023ontology}
Mateiu, P., Groza, A.: {Ontology engineering with large language models}. arXiv preprint arXiv:2307.16699  (2023)

\bibitem{meyer2023llm}
Meyer, L.P., Stadler, C., Frey, J., Radtke, N., Junghanns, K., Meissner, R., Dziwis, G., Bulert, K., Martin, M.: {Llm-assisted knowledge graph engineering: Experiments with chatgpt}. arXiv preprint arXiv:2307.06917  (2023)

\bibitem{peroni2016samod}
Peroni, S.: {A simplified agile methodology for ontology development}. In: OWL: Experiences and Directions--Reasoner Evaluation, pp. 55--69. Springer (2016)

\bibitem{pinto2004diligent}
Pinto, H.S., Staab, S., Tempich, C.: {DILIGENT: Towards a fine-grained methodology for DIstributed, Loosely-controlled and evolvInG Engineering of oNTologies}. In: ECAI. vol.~16, p.~393. Citeseer (2004)

\bibitem{qiang2023agent}
Qiang, Z., Wang, W., Taylor, K.: {Agent-OM: Leveraging Large Language Models for Ontology Matching}. arXiv preprint arXiv:2312.00326  (2023)

\bibitem{simperl2014collaborative}
Simperl, E., Luczak-R{\"o}sch, M.: {Collaborative ontology engineering: a survey}. The Knowledge Engineering Review  \textbf{29}(1),  101--131 (2014)

\bibitem{suarez2012neon}
Su{\'a}rez-Figueroa, M.C., G{\'o}mez-P{\'e}rez, A., Fern{\'a}ndez-L{\'o}pez, M.: {The NeOn methodology for ontology engineering}. In: Ontology engineering in a networked world, pp. 9--34. Springer (2012)

\bibitem{uschold1995towards}
Uschold, M., King, M.: {Towards a methodology for building ontologies}. Citeseer (1995)

\bibitem{viswanathan-et-al-2023-few-shot}
Viswanathan, V., Gashteovski, K., Lawrence, C., Wu, T., Neubig, G.: {Large Language Models Enable Few-Shot Clustering}. arXiv preprint arXiv:2307.00524  (2023)

\bibitem{vrandecic-krotzsch-2014-wikidata}
Vrande\v{c}i\'{c}, D., Kr\"{o}tzsch, M.: {Wikidata: a free collaborative knowledgebase}. Commun. ACM  \textbf{57}(10),  78–85 (sep 2014). \doi{10.1145/2629489}, \url{https://doi.org/10.1145/2629489}

\bibitem{zhang2023llmke}
Zhang, B., Reklos, I., Jain, N., Peñuela, A.M., Simperl, E.: {Using Large Language Models for Knowledge Engineering (LLMKE): A Case Study on Wikidata}. arXiv preprint arXiv:2309.08491  (2023)

\bibitem{zhang-et-al-2023-clusterllm}
Zhang, Y., Wang, Z., Shang, J.: {{ClusterLLM: Large Language Models as a Guide for Text Clustering}}. In: Bouamor, H., Pino, J., Bali, K. (eds.) Proceedings of the 2023 Conference on Empirical Methods in Natural Language Processing. pp. 13903--13920. Association for Computational Linguistics, Singapore (Dec 2023). \doi{10.18653/v1/2023.emnlp-main.858}

\end{thebibliography}

\newpage
\appendix

\section{Survey results}\label{sec:oe-survey}

As outline in Section~\ref{sec:ontochat}, the design of OntoChat relies on our first research question presented in the introduction and recapitulated as follows: \textbf{RQ1} Which ontology engineering activities are the most in need of computational support?

To address RQ1, we run an online survey to collect feedback from ontology engineers.
Based on their experience with OE tasks, they were asked to express their agreement with our statements.
The latter were designed in order to understand the perceived complexity and the need of computational support for OE activities.
Our survey received $N=23$ responses.

\begin{figure}
    \centering
    \includegraphics[width=\linewidth]{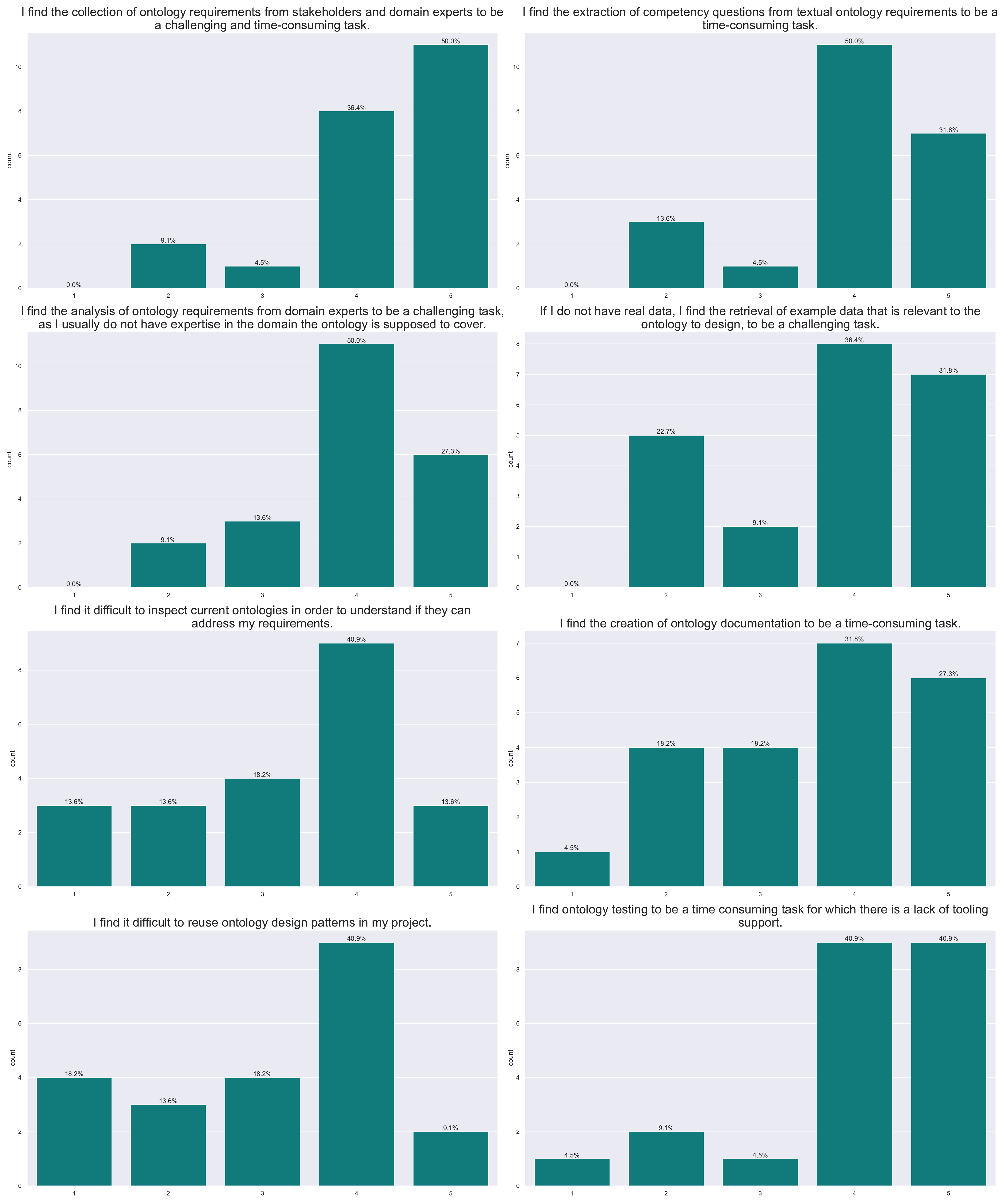}
    \caption{Responses from the Ontology Engineering survey to address RQ1. Replies quantify the agreement of participants with respect to each statement on a 5-point Likert scale, where 1 (absolutely disagree) to 5 (absolutely agree), with 3 being a neutral response (neither agree nor disagree).}
    \label{fig:oe-survey}
\end{figure}

The majority of participants ($60.8\%$) declared to have strong knowledge and expertise in Ontology Engineering (OE).
Participants have formerly used the following OE methodologies:
NEON ($47.4\%$), Ontology 101 ($47.4\%$), eXtreme Design ($42.1\%$), Linked Open Terms ($42.1\%$), METHONDOLOGY ($31.6\%$), and SAMOD ($5.3\%$).
They have experience working an OE projects with ontology design teams of various size:
\textbf{1} ($23.8\%$), \textbf{2-3} ($47.6\%$), \textbf{4-6} ($33.3\%$), and \textbf{7+} ($14.3\%$).
Analogously, they have worked on OE projects involving teams of stakeholders and domain experts of the various size:
\textbf{1} ($15\%$), \textbf{2-3} ($45\%$), \textbf{4-6} ($40\%$), and \textbf{7+} ($25\%$).
The results of this survey are detailed in Fig.~\ref{fig:oe-survey} for consultation.

\section{Evaluation results}\label{sec:apx-eval}

\subsection{Collaborative user story generation}\label{ssec:apx-eval-usg}

\begin{figure}
    \centering
    \includegraphics[width=\linewidth]{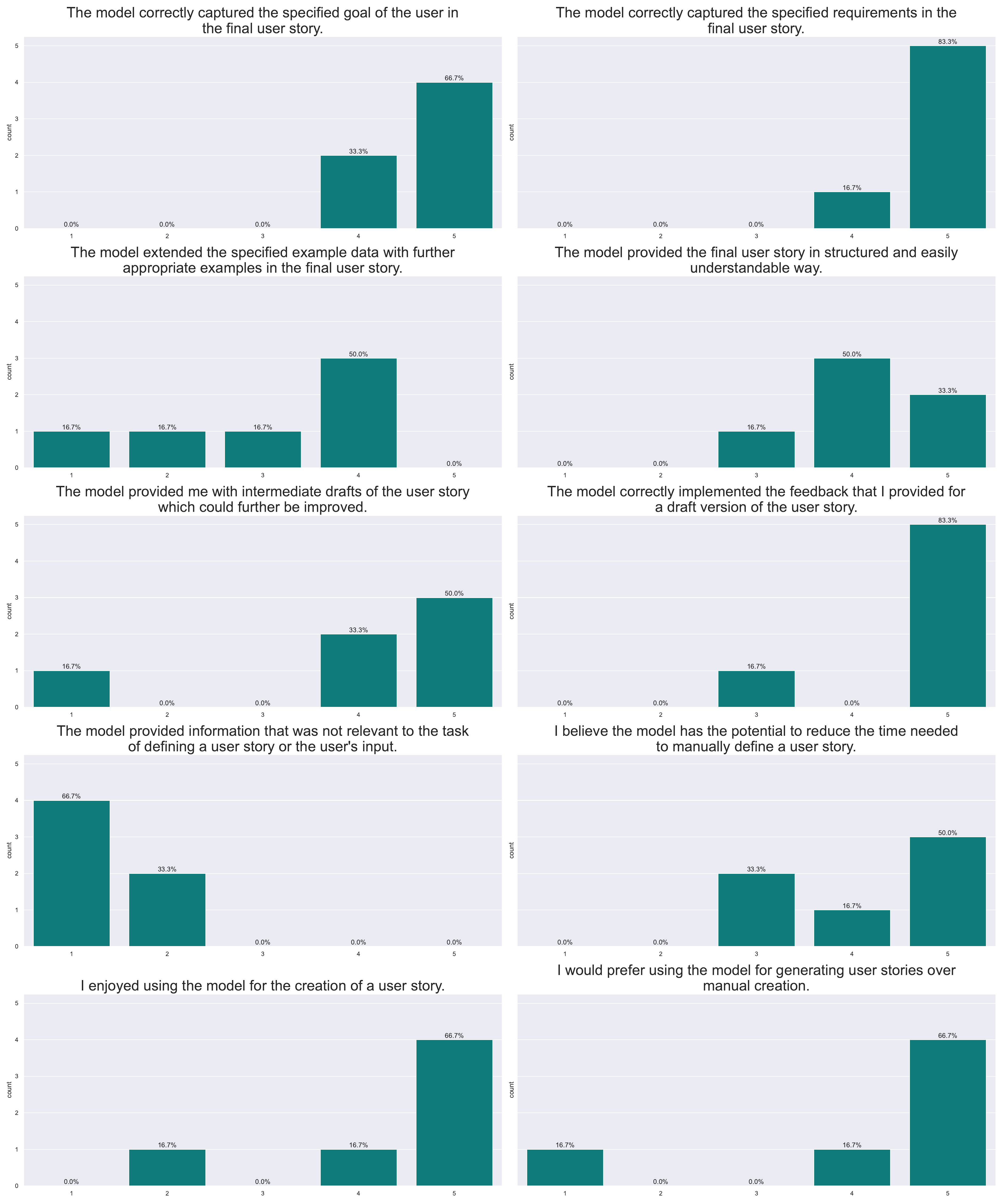}
    \caption{User ratings on the \textit{Collaborative User Story Creation} functionality of OntoChat. The evaluation was performed by $N=6$ domain experts.}
    \label{fig:evaluation-stories}
\end{figure}

This section provides more details on the evaluation of the \textit{collaborative user story generation} feature in OntoChat.
As outlined in Section~\ref{sec:evaluation}, this evaluation was carried out by $N=6$ domain experts who are familiar with the creation of user stories, and were actively involved in the OE activities behind of the Polifonia Ontology Network \cite{polifonia2023pon}. 
In line with our expectations, all participants confirmed their expert knowledge in the music (metadata) domain, and familiarity with the Music Meta Ontology.
In addition, 50\% of them acknowledge \textit{some knowledge} of OE and have experience with the eXtreme Design methodology.
The results of the evaluation are outlined in Fig.~\ref{fig:evaluation-stories} for all the 10 questions.

\subsection{Competency question extraction}\label{ssec:apx-eval-cqe}

\begin{figure}
    \centering
    \includegraphics[width=\linewidth]{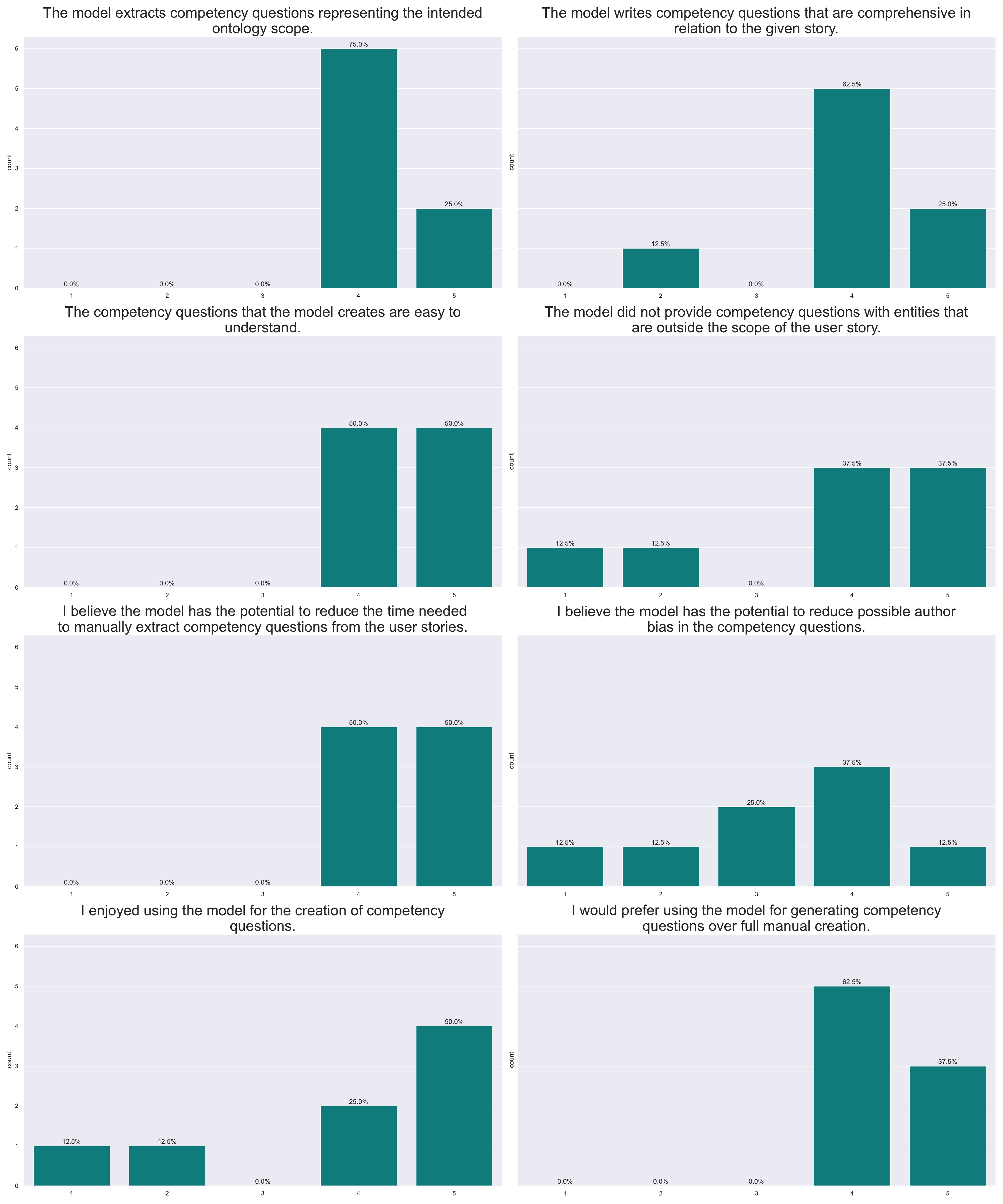}
    \caption{User ratings on the \textit{Competency Question Extraction} functionality of OntoChat. This evaluation was performed by $N=8$ ontology engineers.}
    \label{fig:evaluation-cqe}
\end{figure}

Here, we provide more details on the evaluation of the \textit{competency question extraction} feature in OntoChat.
As outlined in Section~\ref{sec:evaluation}, this evaluation was carried out by $N=8$ participants with expertise in ontology engineering.
The results of the evaluation are outlined in Fig.~\ref{fig:evaluation-cqe} for all the 8 questions.

\subsection{Competency question clustering}\label{ssec:apx-eval-cqc}

\begin{figure}
    \centering
    \includegraphics[width=\linewidth]{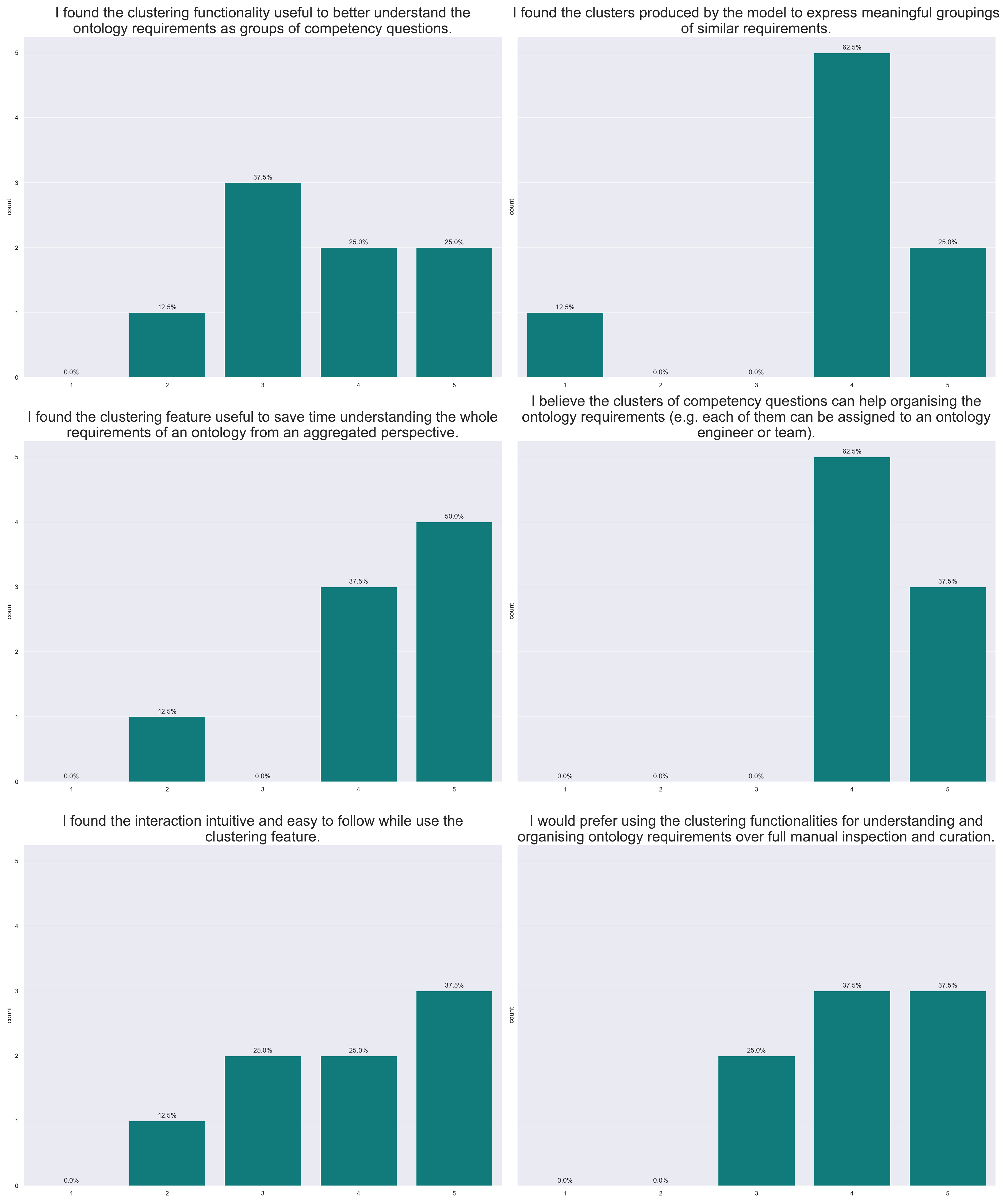}
    \caption{User ratings on the \textit{Competency Question Clustering} functionality performed by $N=8$ ontology engineers.}
    \label{fig:evaluation-cqc}
\end{figure}

As this evaluation step was performed after Competency Question Extraction by the same participants, their background information is the same as reported Appx.~\ref{ssec:apx-eval-cqe}.
The results of the evaluation are outlined in Fig.~\ref{fig:evaluation-cqc}.

\subsection{Preliminary ontology testing}\label{ssec:apx-eval-test}

\begin{figure}
    \centering
    \includegraphics[width=.75\linewidth]{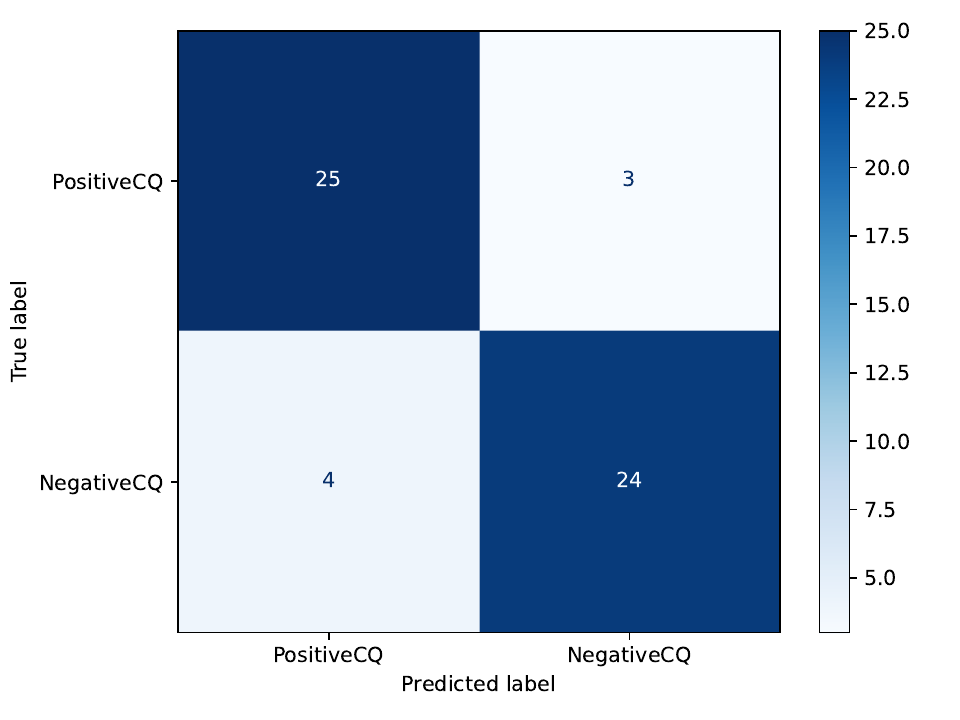}
    \caption{Confusion matrix summarising our results for \textit{prompt-driven CQ unit testing} (c.f. Section~\ref{ssec:ontochat-pretesting}). Positive CQs (target label 1) denote competency questions that are expected to be addressed or covered by the ontology, whereas negative CQs (target label 0) are artificially created to express requirements that are not yet addressed by the ontology and should thus be predicted as such.}
    \label{fig:ontochat-testing-conf}
\end{figure}

To complement our results for ontology testing, we report the confusion matrix in Fig.~\ref{fig:ontochat-testing-conf}, expressing the number of correct predictions (25 true positives, 24 true negatives) and wrongly classified competency questions (3 false positives, 4 false negatives).
Please, note that this can be seen as instance of competency question verification and error provocation for positive and negative CQs, respectively.

\section{Examples}
\subsection{Collaborative user story generation}\label{app-userstory-prompts}
\begin{figure}[ht]
    \centering
    \begin{subfigure}[b]{0.48\textwidth}
        \includegraphics[width=\textwidth]{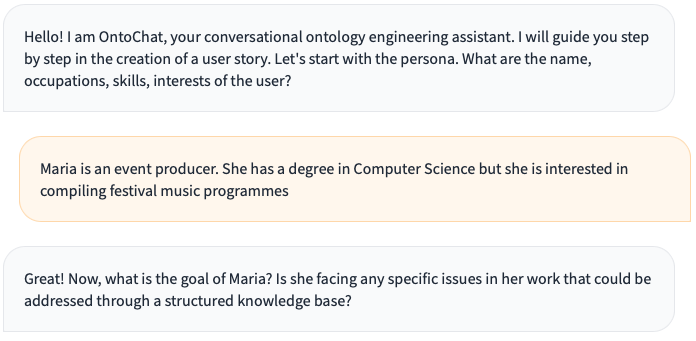}
        \caption{OntoChat collecting information of different story aspects.}
        \label{fig:story-elicitation-steps}
    \end{subfigure}
    \hfill 
    \begin{subfigure}[b]{0.48\textwidth}
        \includegraphics[width=\textwidth]{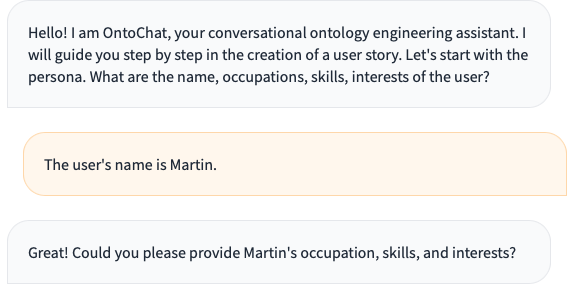}
        \caption{OntoChat dealing with partial answers and missing details,}
        \label{fig:story-elicitaion-partial}
    \end{subfigure}
    \caption{Example excerpts from the knowledge elicitation stage of the user story collaborative generation.}
    \label{fig:story-elicitation}
\end{figure}

\begin{figure}[ht]
    \centering
    \begin{subfigure}[b]{0.48\textwidth}
        \includegraphics[width=\textwidth]{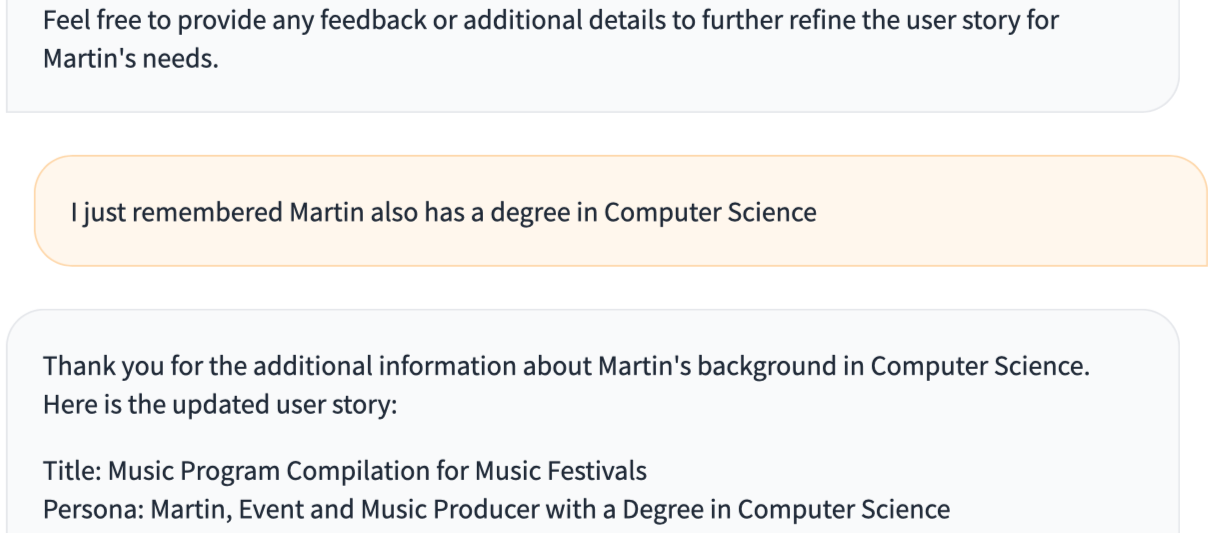}
        \caption{Refinement by additional information.}
        \label{fig:story-refinement-add}
    \end{subfigure}
    \hfill 
    \begin{subfigure}[b]{0.48\textwidth}
        \includegraphics[width=\textwidth]{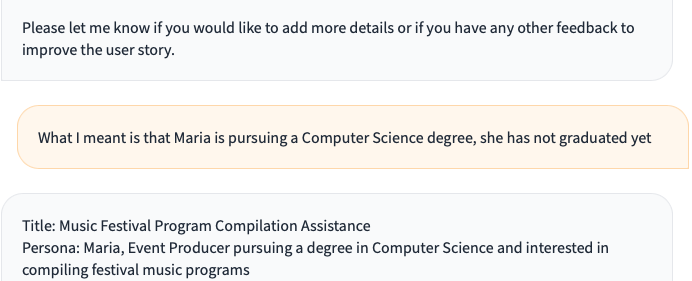}
        \caption{{Refinement through clarification.}}
        \label{fig:story-refinement-clarify}
    \end{subfigure}
    \caption{Example excerpts from the story refinement stage of the user story collaborative generation.}
    \label{fig:story-refinement}
\end{figure}

\begin{figure}[H]
    \centering
    \includegraphics[width=\linewidth]{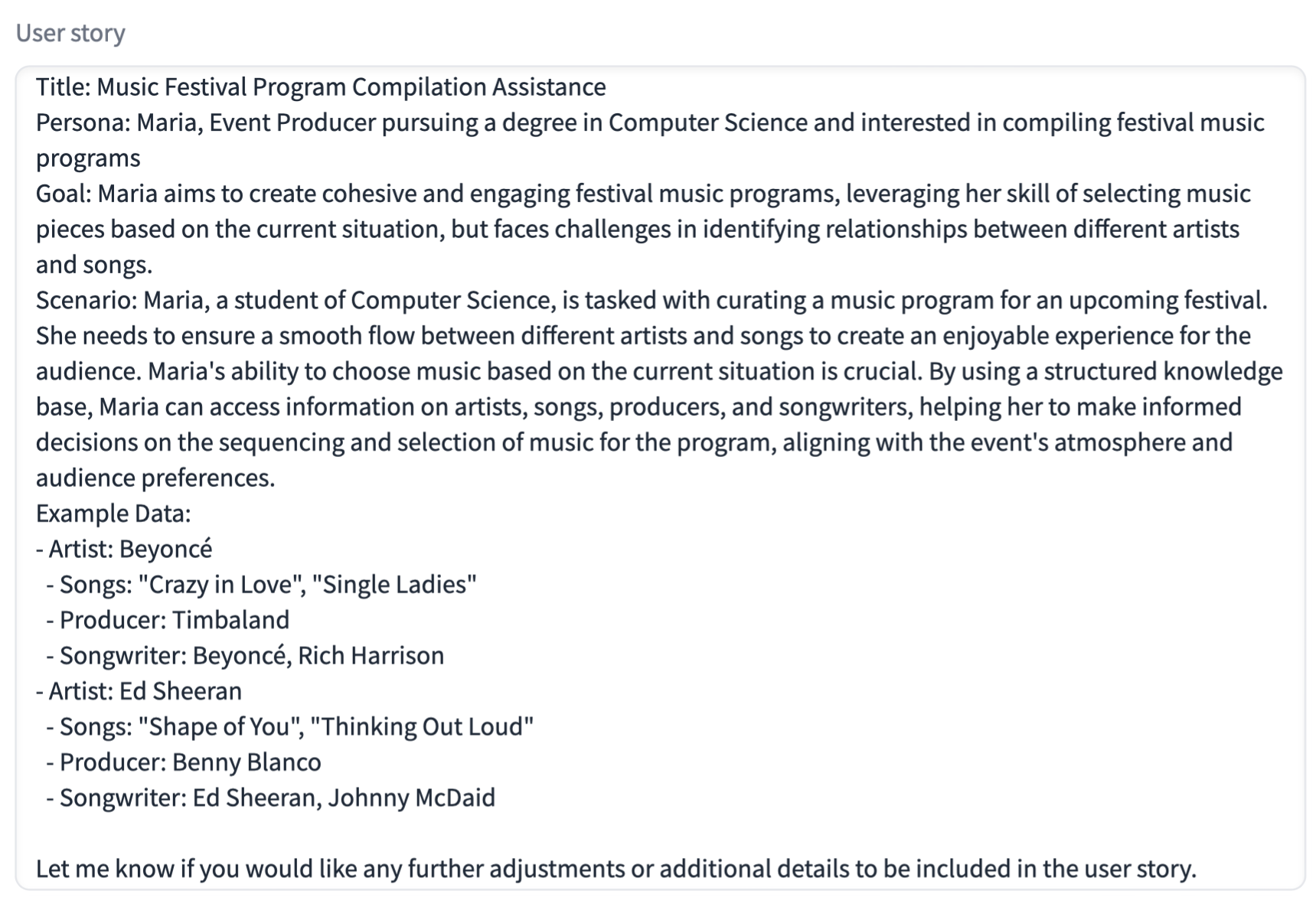}
    \caption{A user story generated with the assistance of OntoChat. }
    \label{fig:story-example}
\end{figure}

In this section, we provide several concrete example excerpts from the user story generation supported by OntoChat. 
Fig.~\ref{fig:story-elicitation} provides outputs of the knowledge elicitation stage while story refinement examples are visualised in Fig.~\ref{fig:story-refinement}. A full user story created with the assistance of OntoChat is shown in Fig.~\ref{fig:story-example}.

\subsection{Competency question extraction}\label{app-cq-prompts}
\begin{figure}[h]
    \centering
    \includegraphics[width=\linewidth]{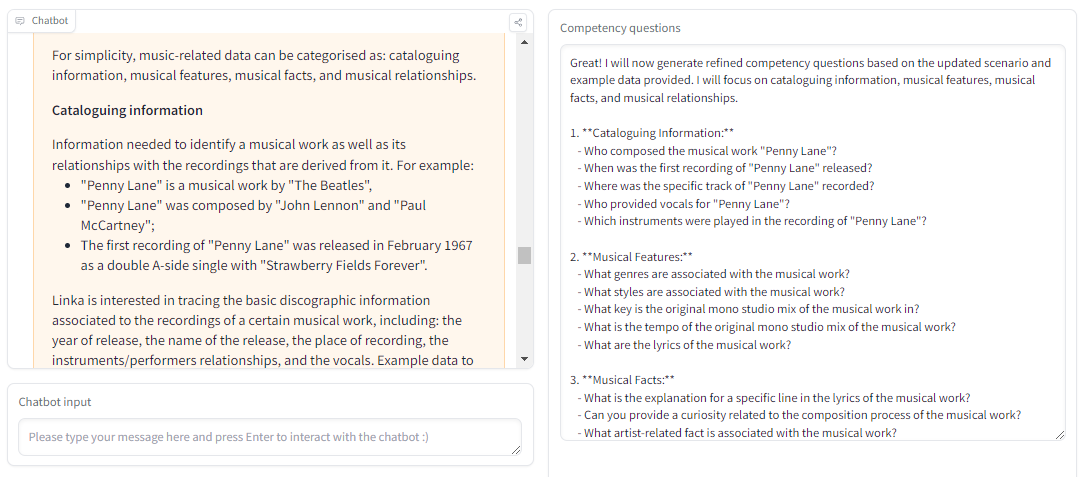}
    \caption{Illustration of OntoChat's \textit{competency question extraction} feature, first extraction of CQs.}
    \label{fig:CQ_first_input}
\end{figure}

\begin{figure}[h]
    \centering
    \includegraphics[width=\linewidth]{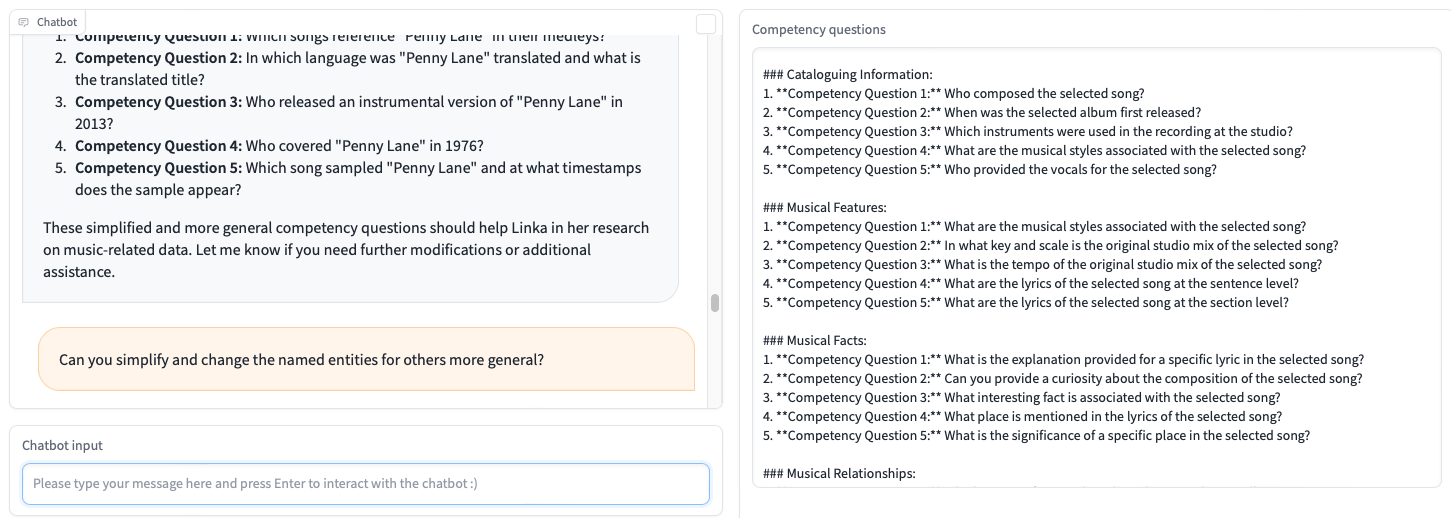}
    \caption{Illustration of OntoChat's \textit{competency question extraction} feature, refinement of the CQs.}
    \label{fig:CQ_refinement}
\end{figure}

In this section, we provide examples from the \textit{Competency Questions Extraction} functionality supported by OntoChat. Fig.~\ref{fig:CQ_first_input} shows the first iteration of CQ extraction from a user story by OntoChat, while Fig.~\ref{fig:CQ_refinement} illustrates the outputs after refinement (c.f. Section~\ref{ssec:ontochat-cqext}). 

\subsection{Competency question clustering}

We report an example of competency question clustering obtained after using this features on a selection of the original CQs driving the design of the Music Meta Ontology \cite{de2023music}, which can be fully found online\footnote{\url{https://github.com/polifonia-project/music-meta-ontology}}.
The clusters produced by OntoChat via the LLM Clustering option (no number of clusters is provided apriori), together with their label, are reported in Listing~\ref{lst:cqc-example}.

\begin{lstlisting}[captionpos=b, caption={Example of CQ clusters generated from Music Meta's (human-annotated) competency questions using the LLM clustering option in OntoChat.}, label={lst:cqc-example}, basicstyle=\scriptsize\ttfamily,frame=single, escapeinside={+}{+}, firstnumber=auto, showstringspaces=false, belowcaptionskip=0em]
"Music Artists": [
    "Which is the name of a music artist?",
    "Which is the alias of a music artist?",
    "Which is the language of the name/alias of a music artist?",
    "Which award was a music artist nominated for?",
    "Which award was received by a music artist?",
    "Which music artists has a music artist been influenced by?",
    "Which music artist has a music artist collaborated with?",
    "Which is the start date of the activity of a music artist?",
    "Which is the end date of the activity of a music artist?"
],
"Musical Pieces and Composers": [
    "Which is the composer of a musical piece?",
    "Is the composer of a musical piece known?",
    "In which time interval did the creation process took place?",
    "Where did the creation process took place?",
    "Which task was executed by a creative action?",
    "Which are the parts of a musical piece?",
    "Which collection is a musical piece member of?"
],
"Music Ensembles": [
    "Which are the members of a music ensemble?",
    "Which role a music artist played within a music ensemble?",
    "Where was a music ensemble formed?"
],
"Musical Performances and Recordings": [
    "Where was a musical piece performed?",
    "When was a musical piece performed?",
    "Which music artists took part to a musical performance?",
    "Which is the recording process that recorded a musical performance?",
    "Which is the recording produced by a recording process?"
], ...
\end{lstlisting}
\suppressfloats[t]




\subsection{Ontology testing}\label{ssec:testing}

We provide a selection of tested competency questions, with expected and predicted outcome (Yes = 1 for supported CQs, and No = 0 for CQs that are not supported by the current definition of the Music Meta ontology).
These are reported in Table~\ref{tab:example-test}, and are complemented by a supporting explanation informing the user on how the requirement can/cannot be addressed by the ontology (e.g. classes and relations that can be used.

\begin{table}[ht]
\centering
\resizebox{\textwidth}{!}{%
{\renewcommand{\arraystretch}{1.4}
\begin{tabular}{lp{12cm}}
\hline
\textbf{CQ} & Which award was received by a music artist? \\
\textbf{Test prediction} & Yes \\
\textbf{Expected test outcome} & Yes (supported) \\
\textbf{Explanation} & The ontology includes the class "Award" and the relation "Received award" which connects a MusicArtist to the Award they have received. This allows for querying the ontology to find the award(s) received by a specific music artist. \\ \hline
\textbf{CQ} & In which time interval did the creation process took place? \\
\textbf{Test outcome} &  \\
\textbf{Expected test outcome} & Yes (supported) \\
\textbf{Explanation} & The ontology described can address the competency question "In which time interval did the creation process take place?" The ontology includes classes and properties related to the creative process, such as CreativeProcess, CreativeAction, and AnnotationMethod. It also includes properties like "isDerivedFrom" and "Creates" that can be used to track the creation process and infer the time interval in which it took place. \\ \hline
\textbf{CQ} & Which is the recording process that recorded a musical performance? \\
\textbf{Test prediction} & Yes \\
\textbf{Expected test outcome} & Yes (supported) \\
\textbf{Explanation} & The ontology can address the competency question "Which is the recording process that recorded a musical performance?" The ontology includes the class "RecordingProcess" and the property "isRecordedBy" which connects a RecordingProcess to a MusicEntity or MusicalPerformance. \\ \hline
\textbf{CQ} & Does a music algorithm favor a specific genre? \\
\textbf{Test prediction} &  Yes \\
\textbf{Expected test outcome} & No (not supported, but potentially plausible) \\
\textbf{Explanation} & The ontology includes the necessary classes and relations to represent music algorithms, music genres, and the relationship between them. By using the "isTrainedOn" and "isTrainedBy" relations, it is possible to infer whether a music algorithm has been trained on a specific music genre or if a music genre has influenced the training of a music algorithm. \\ \hline
\textbf{CQ} & Is a music work associated to any case of plagiarism? \\
\textbf{Test prediction} &  No \\
\textbf{Expected test outcome} & No (not supported) \\
\textbf{Explanation} & The ontology focuses on describing music metadata related to artists, compositions, performances, and recordings, but it does not include specific concepts or relations related to plagiarism detection or identifying cases of plagiarism in music works. To address the competency question, additional concepts and relations related to plagiarism detection and analysis would need to be added to the ontology. \\ \hline
\textbf{CQ} & Which language is most used in a music artist's lyrics? \\
\textbf{Test prediction} &  No \\
\textbf{Expected test outcome} & No (not supported) \\
\textbf{Explanation} & The ontology described does not include any classes or properties related to language or lyrics analysis. It focuses more on the structural and organisational aspects of music metadata, such as compositions, performances, recordings, and their relationships. To address the question of which language is most used in a music artist's lyrics, additional classes and properties related to language analysis and text processing would need to be included in the ontology. \\ \hline
\textbf{CQ} & When was the album first sold? \\
\textbf{Test prediction} &  Yes \\
\textbf{Expected test outcome} & No (not supported) \\
\textbf{Explanation} & The ontology includes classes and properties related to albums, such as Release, Publication Situation, and Publisher. By using these classes and properties, it is possible to model the publication and release of albums, including the specific situation or context in which they are made available to the public. The ontology can capture information about the timing, location, and associated entities involved in the publication process, allowing for the inference of when the album was first sold. \\ \hline
\end{tabular}%
}}
\caption{Examples of tested competency questions with predicted and expected outcome. These are complemented by a supporting explanation.}
\label{tab:example-test}
\end{table}

\end{document}